\documentclass[sigconf,nonacm]{acmart}

\AtBeginDocument{%
  \providecommand\BibTeX{{%
    \normalfont B\kern-0.5em{\scshape i\kern-0.25em b}\kern-0.8em\TeX}}}

\acmConference[CSE598, Date Intensive Systems for Machine Learning, Project Report]{}

\usepackage{subcaption}
\usepackage{graphicx}
\usepackage{listings, xcolor}
\usepackage{hyperref}

\begin{document}

\title{Transformer Architecture for NetsDB}

\author{Subodh Kamble}
\email{skamble3@asu.edu}
\affiliation{%
  \institution{SCAI}
  \institution{Arizona State University}
  \city{Tempe}
  \state{AZ}
  \country{USA}
}

\author{Kunal Kasodekar}
\email{kkasodek@asu.edu}
\affiliation{%
  \institution{SCAI}
  \institution{Arizona State University}
  \city{Tempe}
  \state{AZ}
  \country{USA}
}


\begin{abstract}
Transformers models have become the backbone of the current state-of-the-art models in language, vision and multimodal domains. These models at their core utilize multi-head self-attention to selectively aggregate context, generating dynamic contextual embeddings and modelling long-range dependencies for a clear contextual understanding. Lixi et al. \cite{zhou2022serving} proposed a method to use relational databases for deploying large-scale deep learning models and created an open-source implementation called NetsDB for the same. We build upon the previous work of these authors by creating an end-to-end implementation of the Encoder part of the transformer for model serving in NetsDB. Specifically, we construct a two-block encoder that includes Multi-Head Attention and its accompanying self-attention mechanism, Layer-Norm, Dropout, FeedForward Layers, and the necessary residual connections. We load out weights from our model for distributed processing, deployment, and efficient inferencing. To prove the efficacy of our implementation, we conduct a comprehensive performance analysis by comparing it with existing implementations in PyTorch, Tensorflow, Flax, and MxNet across key metrics such as inference time and model size. 

\end{abstract}



\maketitle

\section{Introduction}
Transformers have become a prominent deep-learning model paradigm widely affecting various deep learning areas specifically Vision and Natural Language Processing domains \cite{li2020survey}. These models perform a state-of-the-art (SOTA) evaluation on almost every task in DL for example machine translation, image inpainting, and object detection. It makes use of the self-attention mechanism for selective context aggregation to filter out relevant details from a wide-range context. It generates contextual embeddings by using the scaled dot product attention to create dynamic embeddings depending upon the context supplied at each attention block. It can also model long-range dependencies and do subword tokenization to effectively have a clear, concise, and focused understanding
of the context at any stage of the transformer block. 

In the last few years, the scale and complexity of deep learning models have exploded. Especially models such as GPT-3, and GPT-3.5 being large-scale language models (LLM) require a distributed cluster for inference/deployment through model and data parallelism \cite{brown2020language}. These models have huge computation overhead and storage costs associated with them resulting in millions of dollars in deployment costs. 

Historically relational databases provide optimized memory locality, even if the working set memory size exceeds the memory capacity through effective buffer pool management \cite{zhou2022serving}. It also provides granular data management, data independence, views and fine-grained authorization. Thus if we can leverage all these capabilities of a relational database for model serving we can minimize operational costs, induce system simplification and correspondingly remove the latency imposed by the decoupling of model serving and data serving systems. Thus we can serve models in relational databases to improve the storage efficiency of the model and improve the model inference compared to other existing deep learning frameworks. We choose NetsDb as our relational database to serve our model and compare it with existing cutting-edge deep learning frameworks such as Flax, MxNet, PyTorch and Tensorflow based on metrics such as inference latency and Storage Size. Serving these models from NetsDB can significantly reduce the system management overhead by eliminating the need to transfer features to decoupled systems for inference. NetsDB integrates data management and query processing, enabling model serving even when the working set size exceeds available memory. 

\subsection{Why NetsDB?}

A set of models with a similar set of parameters/weights will generate similar predictions at the final layer. Thus when we deploy multiple versions of the same trained model for inference, A/B testing, and experimentation we end up with model redundancy and parameter duplication. Netsdb is an open-source implementation of a distributed database system for model deduplication, managing large-scale datasets, and efficient querying. NetsDb can ingest models with varying input shapes, convert them into tensor blocks of the same size, and finally cluster these tensor blocks based on the model accuracy threshold t. Once we obtain a set of similar clusters we use LSH to detect the cluster block to replace based on the similarity measure. This block will be sent to the shared memory and the other blocks will be sent to the private memory.

Further, we can use NetsDB to minimize storage size by packing distinct tensor blocks to optimize the storage space required by Tensor Blocks. This is a variant of the set basis problem wherein we find a subset of vectors that can be used to represent all other vectors in linear space. The authors extend this problem by adding a constraint on each tensor set size in the form of a page size limit. An equivalent class is used so that tensors belonging to the same set are assigned the same class. The authors then use two-stage packing where firstly, they pack tensor blocks in each equivalent class to pages respectively and finally repack the tensor blocks from non-full pages. NetsDB also provides synergistic storage optimization-based model deduplication, page packaging, and caching resulting in reduced parameter size, decreased redundancy through deduplication, improved model efficiency, and reduced inference latency. 

\subsection{Background on the Transformer Encoder Architecture}

\begin{figure}
\centering
\includegraphics[scale=0.5]{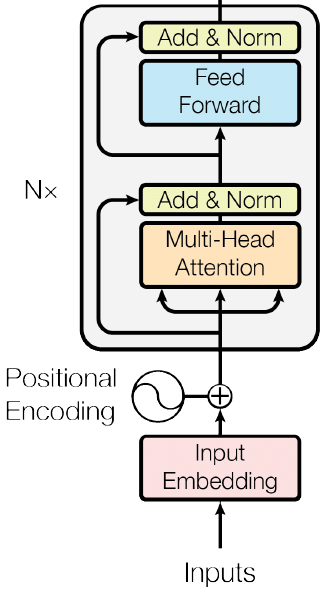}

\caption{Encoder Block}
\end{figure}

The Self-Attention Mechanism uses a scaled dot product attention to selectively aggregate context and filters out relevant details from a wide-range context \cite{niu2021review}. This mechanism generates contextual embeddings that can dynamically adjust based on the context supplied at each attention block. It can also model long-range dependencies and do subword tokenization to effectively have a clear understanding of the context at any stage of the transformer block. Initially, an input sentence in the form of Batches x Sequence Length x Embeddign Size is passed through the Transformer Encoder Block. The input sequence dimension is reshaped in the form of Batches x Heads x SEQLEN x HeadDim and passed onto the self-attention block. In the self-attention block the sequence is passed through linear layers in the form of Wk, Wq, and Wv weights to obtain our k,q, and v pairs. Then we dot our queries, keys, and values to calculate the attention scores. Finally, we apply softmax and multiply with values to obtain the final output. The MHA and Encoder apply the multi-head attention and feedforward neural network sequentially to the input. It also includes normalization and dropout layers to prevent overfitting and corresponding residual layers for feature propagation.

\begin{figure}
\centering
\includegraphics[scale=0.5]{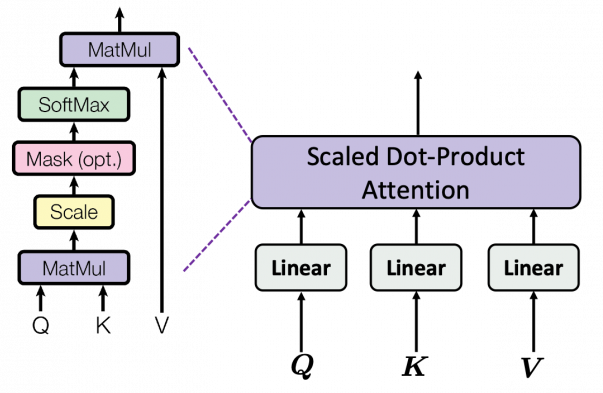}

\caption{Self-Attention}
\end{figure}

\section{Related Work}

\begin{figure}
\centering
\includegraphics[scale=0.35]{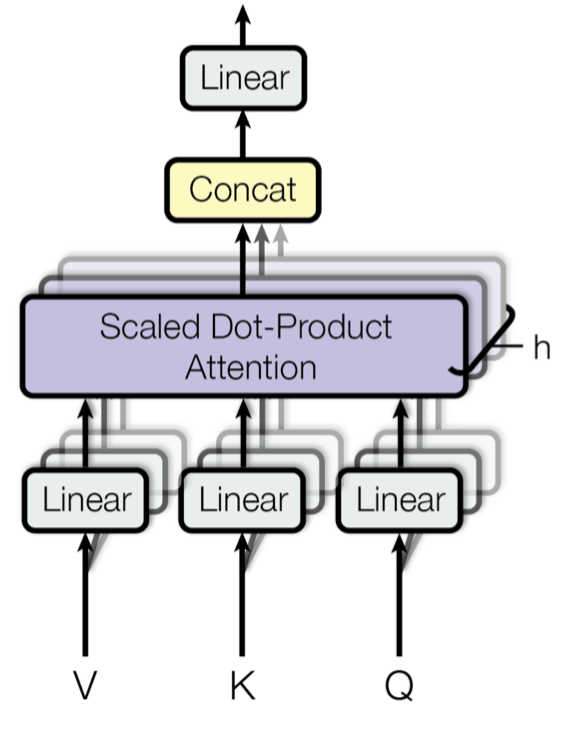}

\caption{Multi-Head Attention}
\end{figure}

For many decades relational databases have been used extensively for managing and storing structured data. Employing these databases for the purpose of building and training machine learning models has recently been on the rise. In this section, we will be reviewing and understanding some of the works of literature published in the field of research.

In \cite{yu2014towards}, Yu K. et al.  proposed a technique for feature selection in big data systems with the combination of relational databases and online learn- ing. The technique has achieved high-performance metrics on several benchmark datasets. The technique also handles categorical and numerical features of datasets with scalable performance using the algorithm of online feature selection.

Lixi et al. \cite{zhou2022serving} proposed a method to use relational databases for deploying large-scale deep learning models with model de-duplication to identify shared parameters among multiple model versions and reduce model size via shared memory. Using a two-step procedure that requires training the model on massively high-volume datasets and deploying it by storing the parameters in a relational database with deduplication, it is able to significantly reduce the computational overheads and storage costs associated with it. Their implementation in the form of NetsDb makes use of multiple storage optimization techniques such as caching, page-packing, and duplication detection. This in turn results in enhanced and optimized systems for large-scale model optimization and serving in database.

Binhang Yuan et al. \cite{yuan2021tensor} introduce a novel abstraction for handling large-scale models and scaling them beyond a single machine. Tensor Relational Algebra proposed by the authors makes use of tensor operations to auto-optimize and make them highly scalable due to their parallelizable nature. Computation expressed as TRA is converted into its native implementation algebra which makes it highly optimizable and fast. The authors claim that the TRA-based backend easily outperforms competing ML workflows running on distributed clusters. TRA has a limited set of applications and requires significant modifications for effective adaption into different implementation domains (Requires domain-specific knowledge). Storing sparse tensors effectively in an optimal way via high-performance kernels can be an issue.

Peter Bloem [6] in his article titled: "Transformers from scratch", introduces key concepts in building self-attention modules from scratch such as self-attention, its key role in selective context aggregation, the significance of Query, Key, value pairs and their corresponding weights. By Building upon the granular understanding of the aforementioned concepts the author builds up the transformer block from scratch by sequentially building up the self-attention layer, layer normalization, feedforward layer corresponding residual connections and multihead attention \cite{vaswani2017attention}. The author provides code blocks for each implementation for elucidating the understanding and implementation workflow for transformer blocks from scratch.

\section{Experimental Setup}

Setting up the environment for NetsDB was a challenging task that required perusing the source code and multiple days of debugging to resolve build errors, and jumping to and fro between various environment setups. NetsDB is built upon an earlier version of an open-source project called PlinyCompute. After experimenting with Virtual Box, Docker, AWS, and Agave Compute (ASU's Supercomputer), we were finally able to set up NetsDB successfully by building it through scons on Agave using a singularity container.
To achieve this, we built a singularity definition file with the required dependencies and utilized a compute node with 16 CPU cores and sufficient RAM and storage to run the singularity container with NetsDB. We built all the dependencies, including our multi-head attention module, using scons, started the pseudo cluster, and ran our binary file generated from the compilation of scons.

\subsection{Overall Issues faced}
\begin{itemize}
     
\item Our attempt to execute the scons command based on the instructions provided in the README file was unsuccessful due to the requirement for clang++ version 17, which was not available on our Ubuntu system that only had version 10.
\item To address this issue, we opted to use a Virtual Machine running Ubuntu 22.04 to compile NetsDB, but we encountered lengthy processing time when running the scons command.
\item Another approach we tried was to use an AMI file containing pre-installed packages required for running NetsDB. We created a t2 micro instance from this image, but after waiting 3-4 hours for the compilation to complete, we realized that we needed a larger instance with more memory and SSD. We then switched to using an r4.large instance, which resulted in a successful compilation. However, we faced a challenge when attempting to expose the localhost:8009 port used by NetsDB to interact with other processes. We tried adding the address to Ubuntu using the iptables command and explored other options, but we were unable to run the binary file generated from the scons command.

\item We faced issues when using some predefined functions. To overcome this issue we checked the usage of the namespace keyword, and if the error persists then we copied that slice of code into the destination file. 

\item We faced an issue while writing our own libraries regarding which computation to inherit, for example, joincomp and selectcomp. To overcome this issue, we attempted to understand each different class and then determined which one would be the most useful. One suggestion that may be helpful is to add more descriptive comments in the code.

\item To develop our own functionalities, we reviewed some existing ones to gain insight. During this process, we overrode two methods: getprojection and getselection. However, we encountered difficulties while implementing getselection due to the absence of documentation for these functions. With the guidance of our teaching assistant, we were able to comprehend the underlying logic behind these methods.
\item After the successful compilation of the Scons command, the shared library ".so" files were not present in the build folder. This eventually caused an error while running the binary test file. To overcome this issue, we had to follow a specific naming convention for compiling the shared libraries (layernorm, resconnection) so that it generated the ".so" files in the build folder.
\item Due to insufficient documentation for SCons, we faced some issues while compiling our own shared libraries and transforming our test file into binary. We tried to understand the SCons file and how it behaves. After trial and error, we found that an additional argument was needed to import shared functionality in the SConstruct's program API.

\end{itemize}

\subsection{Singularity}

We suspected that lack of memory and computing power may have been the culprit. In spite of our multiple attempts to debug the issues and get the aforementioned solutions working we were unsuccessful. To alleviate the issues of lack of computing resources we moved our codebase to ASU's Agave Supercomputer. Agave requires computations to be carried on through an interactive session or SBATCH script on the login node. We found that an interactive session with a single compute node and 16 CPU Cores fulfilled our purpose. Docker is incompatible with HPC resources as it requires a root user, hence we use singularity as a replacement. We create a definition file with all the dependencies outlined in the NetsDB documentation and build a singularity container to create an isolated, reproducible and portable environment for experimentation. We had the following issues with Singularity:

\begin{itemize}
\item Singularity does not permit the user to install packages after spinning up the container. Making sure the multitude of packages has been installed requires multiple iterations of updating the def file and building the image - a redundant effort.
\item There were incompatibilities between various singularity modules and we were able to narrow it down to 3.8.0 as the compatible module.
\item Compared to Docker, Singularity suffered from bloating due to unused packages and dependencies from NetsDB impacting performance and storage. 
\item The compute node required a CPU with at least 8 cores, ample RAM and Storage for the Pseudo Cluster to start, and the Binary file generated by scons to run without any issues.

\end{itemize}

\section{Methodology}
\subsection{Systems for Comparison}

We are making use of NetsDB only for inference and PyTorch \cite{paszke2019pytorch}, Tensorflow \cite{abadi2016tensorflow}, MxNet \cite{chen2015mxnet} and Flax for both Inference and Training. We do this to make a fair comparison between a relational deep learning model serving platforms such as NetsDb and popular deep learning frameworks. In addition to the popular deep learning frameworks PyTorch and Tensorflow, we also utilized Flax and MxNet for our implementation. Flax is based on Jax with XLA and improved Autograd while MxNet is utilized for distributed training and deployment of large-scale machine learning models.

\subsection{General workflow across all the frameworks}

Our PyTorch Implementation primarily consists of a Transformer Encoder with sequential Multi-Head Attention (MHA)layer, Feedforward layer, layer normalization and dropout. In the forward block, we apply prenorm, MHA, followed by dropout and corresponding dropout layers for our first encoder block. We have two such encoder blocks. Our sequence length is 10 with a batch size of 2 and embedding dimensions of 64 and 4 heads. We pass random sentences for inference in the form of Batch x Sequence x EmbeddingDim. Currently, our focus is only on implementing the encoder block i.e. a model-centric approach opposed to an accuracy-centric approach. Thus we implement and test our models on untrained architectures as we want to evaluate the inference latency, and model size rather than the accuracy at this stage of our project. Coming to multi-head attention, we take the attention heads, output dim and dropout probability as input. Once passed the MHA block, firstly we apply 3 linear layers which are Wk, Wq and Wv computing the K (Key), Q (Query) and V (Value) matrice values. Further, we split it into multiple heads using a reshape-based splitHeads() function. We then apply attention to the projected K, Q and V values to get the attention scores and then unify the heads through a linear layer. To get the attention scores the transposed K is multiplied with Q, scaled via the square root of the dimension of K. Finally we apply softmax to these attention scores and multiply it with the values matrix to obtain to get our self-attention values.

\subsection{PseudoCode for Encoder Block}
\begin{lstlisting}[showstringspaces=false, frame=single, mathescape=true]
function selfattention(k,q,v){
    scores = $q \ x \ k^t$/root(dimension(k))
    scores = softmax(scores)
    out = scores x values
    Return out
}

function multihead(tensor:t){
    k = $W_k(t)$
    q = $W_q(t)$
    v = $W_v(t)$
    k,q,v = splitheads(k,q,v)
    scores = selfattention(k,q,v)
    out = self.unify_heads(scores)
    Return out
}

function Encoder(tensor:t){
    [ti = prenorm(t)
    t = ti + dropout(multihead((t)))
    ti = prenorm(t)
    t = ti + dropout(multihead((t)))]x2
    return t
}
\end{lstlisting}

\subsection{Implementation of Transformer Block in NetsDB}

\begin{figure}[!htbp]
\centering
\includegraphics[scale=0.3
]{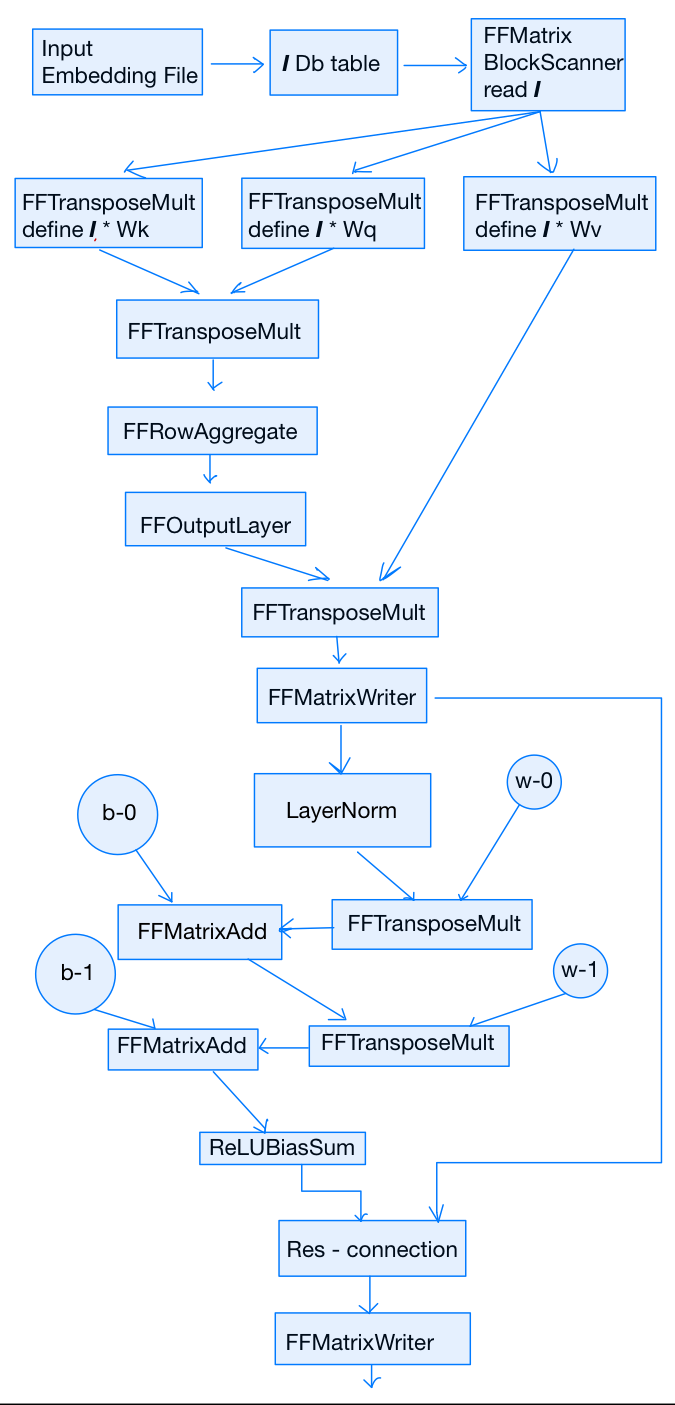}

\caption{Flowchart}
\end{figure}

The Solution Architecture consists of the following components:
\begin{itemize}
    \item Input files (.out files) or randomly generated values.
    \item Weights and biases files for each key, query, and value feed-forward network.
    \item An input vector that specifies the context length of subwords.
    \item A Pliny DB (PDB) client for establishing a connection with the database, PDB computation for performing database operations, and a PDB handle for handling DB typecasts.
    \item We read and store data in the sets of the FFMatrixblock before computing them in the PDB computation environment.
\end{itemize}

Now coming to our model we constructed a transformer encoder block consisting of four essential building blocks, namely the attention mechanism, feed-forwarded neural network, normalization, and residual connections. To build the block, we first needed a separate folder structure containing header and source files for building a transformer block. We created two folders, one for LayerNormalization and another for ResidualConnections, with the header folder defining the supporting functions and the src folder containing the necessary implementation for that particular operation. As the LayerNormalization and ResidualConnections operations were not built previously, we implemented them from scratch.

For LayerNormalization, we took one matrix block on which we wanted to perform normalization and calculated the mean, variation, and standard deviation for each separate batch size of data. After reading the block and performing the necessary operations, we updated the values in the result matrix block and returned them. Note that the two input blocks and an output block will have the same dimension. Similarly, for ResidualConnections, we used the JoinComputation to perform elementwise matrix addition after calculating the number of 2D blocks. Here again, the two input blocks and an output block will have the same dimension.

For the feedforward network, we opted for two hidden layers and derived them from the already present functionalities such as matrix multiplication, matrix transposition, and Relu activation function. The main crux of the transformer block is the attention mechanism, and we focused on transforming the input vector across the key, query, and value matrices. We transposed the Key matrix and multiplied it elementwise with a Query vector. We divided the softmax operation into two parts, scanning all the row elements of every block from the input and computing the exponential for each value in the first part, and aggregating the exponential values in the next part and dividing them by the total value. We stored the updated values in another result matrix block and then returned them, writing them back to the database using the intermediate writer. 

We then performed normalization on the resulting output using the Layer normalization operation to normalize the data by the batch size. This normalized output was stored in the database via the intermediate writer and served as input for the Feed-forward network. This input set passed through the first hidden layer and underwent MatrixBlock multiplication with a w0 set of the weight matrix. The resulting set of blocks underwent MatrixBlock addition with a b0 set of the bias matrix, followed by the ReLu activation. As the network had two hidden layers, we repeated this process with a w1 set of the weight matrix and a b1 set of the bias matrix and performed a final ReLu activation. The resulting output set was then forwarded to the Residual connection operation along with the output set we got from the layer normalization operation. The resultant matrix we got from the residual connection operation is the output of one transformer block.
To complete this transformer task, we required a total of seven sets of weight files, with the attention mechanism taking key, query, and value matrix as trained parameters during inference time, and the feed-forward network requiring two sets of weight matrices and two sets of bias matrices as the network contains two hidden layers.

\begin{table}[!htbp]
    \centering
    \begin{tabular}{ |p{2cm}|p{4.5cm}|p{1cm}| } 
 \hline
     Filename  & Gist about the file & Lines of code \\ 
 \hline
 FeedForwardNet & Implementation of 2 hidden layers FFN  & 213\\ 
 MHATest & Logic of attention mechanism and testing for the full transformer block & 482\\ 
 LayerNorm & Implementation of Layer Normalization & 96\\
 ResConnection & Implementation of Residual connection & 149\\ 
 \hline
\end{tabular}
\\
    \caption{Files added in the existing NetsDB implementation}
    \label{tab:Table 1}
\end{table}

\subsection{Other comparative Implementation}

    \subsubsection{PyTorch}
    The Torch library, a numerical computing library that supports tensor operations, is the foundation upon which PyTorch is constructed. PyTorch performs forward propagation using the computational graph at inference time. This entails applying the weights and biases of each layer to the input data before computing the layer's output. The input data is then passed through the model layer by layer. Up until the final output of the model is generated, each layer's output serves as the input for the subsequent layer. Additionally, PyTorch offers GPU acceleration, which can greatly speed up the computations involved at inference time. Moreover, it can be seen that it outperforms NetsDB and Tensorflow by a significant margin.
    \subsubsection{TensorFlow}
    TensorFlow internally represents the computations made by the model via a computational graph.TensorFlow can optimize the computational graph prior to execution thanks to the lazy execution methodology. Running a TensorFlow application doesn't actually involve any calculations. Instead, it creates an abstract representation of the computations to be done called a computational graph. Forward propagation is carried out by TensorFlow using the optimized computational graph at inference time. TensorFlow leverages its lazy execution model during forward propagation to only compute the portions of the computational graph required to compute the final output. This can result in significant processing and memory savings, especially for complex, large-scale models. Although TensorFlow is quite popular amongst ML researchers, the result shows that it performs remarkably slow than PyTorch and flax. 
    \subsubsection{Flax}
    
Flax is a neural network library that utilizes JAX under the hood. JAX is an improved and accelerated version of NumPy with XLA and enhanced Autograd capabilities. With its accelerated backend, Flax offers high performance and flexibility in addition to ensuring safety, control, terse code, and highly functional APIs. By exposing the full power of JAX through XLA, it achieves the best performance in terms of average inference time.

However, it is worth noting that writing the Encoder block in Flax can be convoluted and requires more effort due to a lack of developer support and non-extensive documentation. We encountered frequent errors during the process and spent a significant amount of time debugging. Surprisingly, the model size is the second-largest, just behind NetsDB, as it takes a whopping 518 KB.

\subsubsection{MxNet}
MxNet is a flexible and efficient deep-learning framework that empowers the programmer to mix imperative and symbolic approaches, resulting in maximized productivity, portability, and scalability. Thanks to these features, MxNet achieves almost the best inference time compared to other frameworks while maintaining a relatively small model size.

However, MxNet shares a common issue with Flax, which is the lack of community support, documentation, and the need for extensive debugging when implementing the Encoder block.

\section{Results}
\begin{table}[!htbp]
    \centering
    \begin{tabular}{ |p{2cm}|p{3cm}|p{3cm}| } 
 \hline
     Framework  & Average Inference Time (ms) & Average Model Size (KB) \\ 
 \hline
 NetsDB & 9.7634 (seconds) & 52 (MB)\\ 
 PyTorch & 127.3798
 & 135\\
 Tensorflow & 831.5501 & 154\\ 
 MxNet & 59.5560 & 138\\
 Flax & 26.4957 & 518\\ 
 \hline
\end{tabular}
    \caption{Performance comparison of NetsDb with other deep learning frameworks}
    \label{tab:Table 2}
\end{table}

We have implemented a two-encoder MHA block architecture across multiple frameworks including PyTorch, TensorFlow, MxNet, and Flax for fair performance evaluation against NetsDb. While PyTorch and TensorFlow are common DL frameworks, we also utilized Flax which is built upon Jax with XLA and improved Autograd, and MxNet for distributed training and deployment of large-scale machine learning models. Our implementation includes layer norm, self-attention, feedforward layers, encoder+mha blocks, and residual layers built from scratch in all the frameworks. Additionally, we experimented with instance normalization to ensure consistent statistics across small batches. We have the following observations from the results table:
\begin{itemize}
    \item From the table below we observe that Flax has the fastest inference time clocking at around 26.49 ms, followed by MxNet at 59.55 ms, PyTorch at 127.37 ms, TensorFlow at 831ms and NetsDB slowest at 9.78 ms.
    \item For the model size we observe that the smallest model size is from PyTorch followed by MxNet, Tf, Flax and finally NetsDb with the largest model size of 52 MB.
    \item Although NetsDB may have a larger model size and larger inference latency, it makes it up by uncoupling model serving and storing as well as providing all the benefits of a relational database such as fine-grained control, data management, query processing whilst enabling model serving even when the working set size exceeds available memory
\end{itemize}

\section{Conclusion}

We have taken up the challenging task of implementing the Transformer Encoder block from the ground up in NetsDB, prioritizing optimization and efficiency. With meticulous attention to detail, we craft exhaustive test cases to guarantee the precision of our architecture. To showcase the prowess of our implementation, we conduct a comprehensive performance analysis, pitting it against established implementations in PyTorch, TensorFlow, Flax, and MxNet, using critical metrics such as inference time and model size.

To ensure that our implementation is easily accessible and reproducible, we leveraged Singularity to containerize NetsDB on Agave. This approach will enable us to easily share our implementation with other researchers and make it easy to reproduce our results. Our implementation of the Transformer Encoder block in NetsDb demonstrates the potential of integrating cutting-edge deep learning architectures with popular databases. This could lead to more efficient and effective natural language processing, machine learning, and deep learning applications.

\section{Future Work}

In order to advance our understanding and implementation of NetsDB, we have set forth several future plans. First and foremost, we plan to continuously refine our UDF implementation and extend it to incorporate the MLP-Mixer Model \cite{tolstikhin2021mlp}. Another objective is to develop a linear and fully-convolutional ViT \cite{dosovitskiy2020image}. Additionally, we aim to conduct further experiments with model deduplication to optimize the efficiency of our system. Our plan is to merge our Encoder Block codebase with asu-cactus NetsDB and subsequently implement the Decoder Block as well.

Finally, we intend to conduct extensive evaluations with additional metrics and diverse datasets to gain deeper insights into the performance and capabilities of NetsDB. By pursuing these goals, we hope to further enhance the effectiveness and efficiency of NetsDB for various Deep learning applications. 

\begin{acks}
This project, report and the research behind it would not have been possible without the support, insights and feedback from Professor Jia Zou and Lixi Zhou.
\end{acks}

\section*{Code}
\textbf{Github Link:} \url{https://github.com/kn0wthing/netsdb}


\bibliographystyle{ACM-Reference-Format}
\bibliography{sample-base}
\end{document}